\documentclass[journal]{IEEEtran}
\usepackage{amssymb}

\usepackage{cite}

\ifCLASSINFOpdf
  \usepackage[pdftex]{graphicx}
\else
\fi
\usepackage{amsmath}
\begin{document}
%
\title{Improving Target-driven Visual Navigation with Attention on 3D Spatial Relationships}
%
%
%

\author{Yunlian~Lv,
	Ning~Xie,
	Yimin~Shi,
	Zijiao~Wang,
	and~Heng~Tao~Shen
\thanks{Yunlian Lv is with Center for Future Media, School of Computer Science and Engineering, University of Electronic Science and Technology of China, Chengdu, China. e-mail: yunlian@std.uestc.edu.cn.}}
\maketitle

\begin{abstract}
Embodied artificial intelligence (AI) tasks shift from tasks focusing on internet images to active settings involving embodied agents that perceive and act within 3D environments. In this paper, we investigate the target-driven visual navigation using deep reinforcement learning (DRL) in 3D indoor scenes, whose navigation task aims to train an agent that can intelligently make a series of decisions to arrive at a pre-specified target location from any possible starting positions only based on egocentric views. However, most navigation methods currently struggle against several challenging problems, such as data efficiency, automatic obstacle avoidance, and generalization. Generalization problem means that agent does not have the ability to transfer navigation skills learned from previous experience to unseen targets and scenes. To address these issues, we incorporate two designs into classic DRL framework: attention on 3D knowledge graph (KG) and target skill extension (TSE) module. On the one hand, our proposed method combines visual features and 3D spatial representations to learn navigation policy. On the other hand, TSE module is used to generate sub-targets which allow agent to learn from failures. Specifically, our 3D spatial relationships are encoded through recently popular graph convolutional network (GCN). Considering the real world settings, our work also considers open action and adds actionable targets into conventional navigation situations. Those more difficult settings are applied to test whether DRL agent really understand its task, navigating environment, and can carry out reasoning. Our experiments, performed in the \emph{AI2-THOR}, show that our model outperforms the baselines in both SR and SPL metrics, and improves generalization ability across targets and scenes.
\end{abstract}


%
\IEEEpeerreviewmaketitle

\section{Introduction}
From a home service robot asked to ``open the cabinet under the coffee machine and give me a cup inside'' to a device that helps its visually impaired wearer navigate an unfamiliar subway, a wide range of abilities need to be demonstrated for the next generation of AI-powered assistants. To develop these skills, many researchers believe that the most effective way is to focus on embodied AI tasks, such as visual navigation \cite{zhu2017target}, instruction following \cite{anderson2018vision}, and embodied question answering (embodied QA) \cite{das2018embodied}. These tasks ground system's training using interactive environments instead of relying on static datasets (e.g. ImageNet \cite{deng2009imagenet}, COCO \cite{lin2014microsoft}, VQA \cite{krishna2017visual}). Compared with internet image dataset-based tasks, embodied AI tasks require special skills of active perception, long-term planning and learning from interaction.

\begin{figure}[!t]
	\centering
	\includegraphics[width=3.5in]{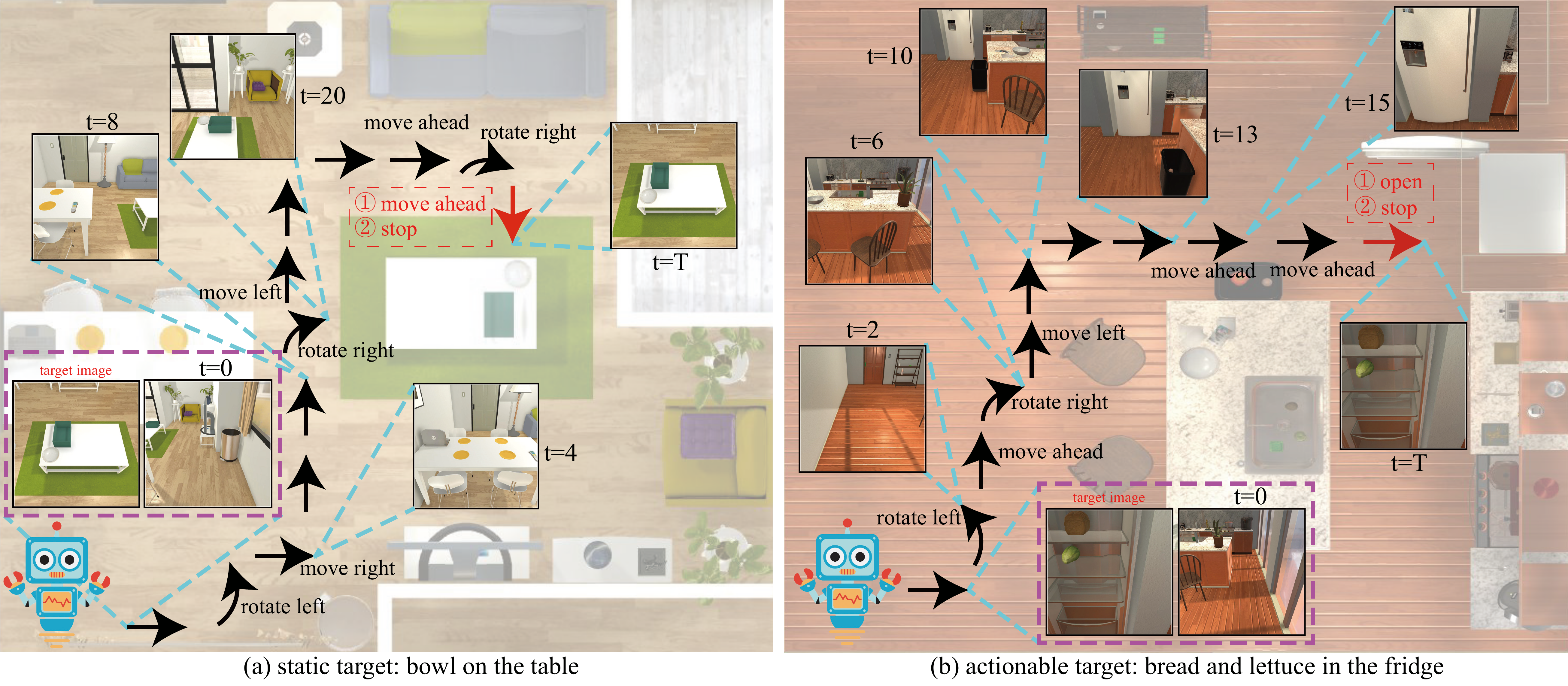}
	\caption{Examples of target-driven visual navigation in 3D indoor environments. The agent takes the current observation and target image as input, and then outputs an action that would reach the target position. Our targets consist of static targets (e.g. the bowl on the desk (a)) and actionable targets (e.g. the bread and lettuce in the fridge (b)). We also show the egocentric view of the agent and its corresponding actions at a few time steps. Red actions represent the last two successful actions to finish the navigation task.}
	\label{fig-target}
\end{figure}

In this paper, we focus on the target-driven visual navigation in 3D indoor scenes, where agent perceives its environments through egocentric views and can perform a series of atomic actions, such as move ahead, look up, or turn right. A natural way to instruct a robot is to ask it to move near a certain location, such as a place (e.g. ``go to sofa") \cite{chaplot2018gated,wortsman2019learning} or someone's room (e.g. ``go to kitchen") \cite{wu2018building}. Besides, a robot can also carry out a natural-language instruction or find useful information to answer a question \cite{anderson2018vision,das2018embodied}. Similar to \cite{zhu2017target}, instead of using language to command a robot, we communicate with the robot by showing it a single image of a faraway target (see Fig. \ref{fig-target}). The agent is required to intelligently navigate to a destination from any possible starting positions according to the assigned target image. At each time step, the agent observes its environment and matches with the given target image meanwhile, then determines its next action.

\begin{figure*}
	\centering
	\includegraphics[width=6.5in]{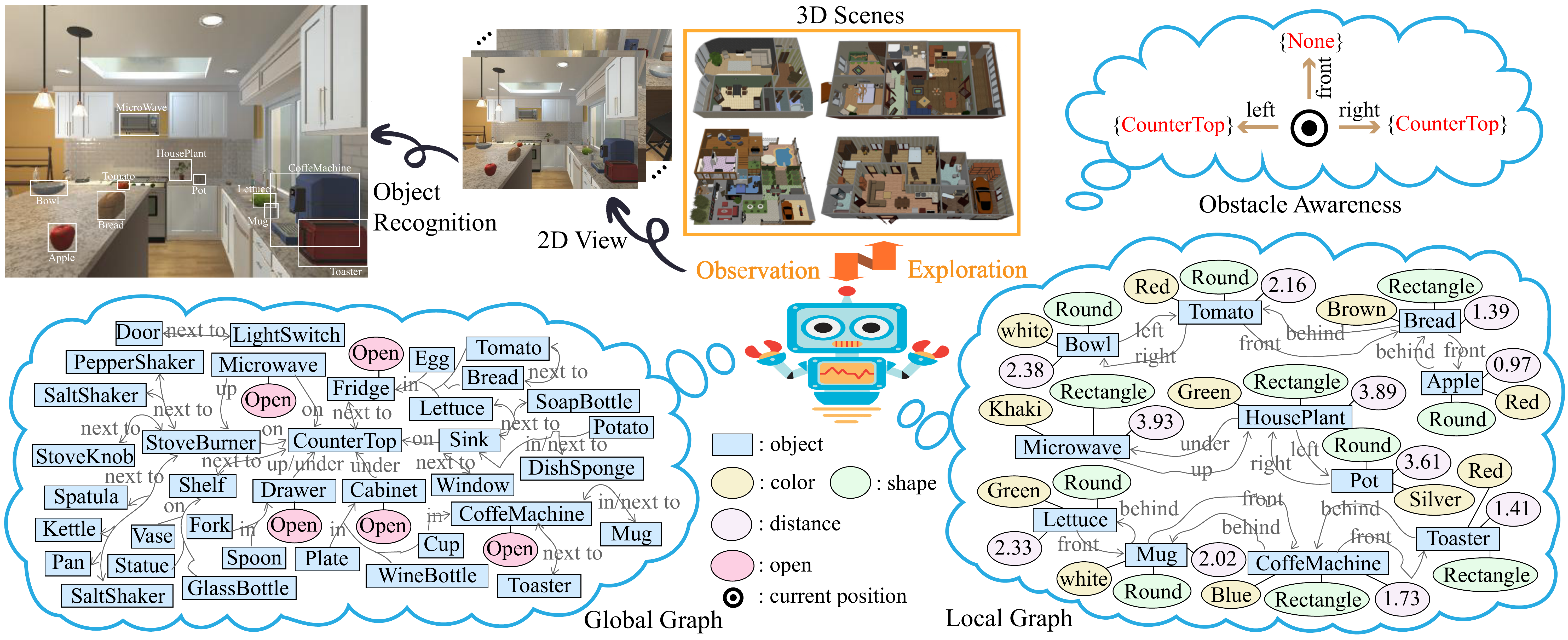}
	\caption{Illustration of powerful 3D knowledge in our visual system, which has at least four indispensable capabilities that enable navigation tasks to be successfully completed. The first is the ability to recognize object attributes from 2D view, including object, color, size, shape, distance, etc. The second is the ability to perceive obstacles, which means that obstacles in several directions can be aware of so as to avoid collisions. The third is the ability to depict the adjacency relationship of objects from the current field of view, which named local graph in the figure. Orientation information (e.g. left/right, front/behind, up/under) between objects can be easily extracted. The last one is the ability to summarize the rules of objects placement after visiting numerous rooms as shown in global graph, which indicates what objects are usually placed beside or inside them and allow to reasoning about possible finding locations. The global graph in our figure simply shows interaction attribute but omits other object's attributes information. These powerful capabilities enable us to navigate without collisions and adapt to various variations in novel and unfamiliar environments.}
	\label{fig-knowledge-graph}
\end{figure*}

Training an agent with intelligent navigation capability like human beings (e.g. few-shot learning, avoid collisions, target-induced exploration) faces lots of difficulties and challenges. Most current approaches usually use pixel-level visual features to learn optimal policy via popular deep reinforcement learning algorithms to overcome these problems. However, agent trained in this way cannot perform well across unseen targets and scenes\cite{dhiman2018critical,wu2018building}. In general, the trained agent can exploit scene information on training targets, yet unable to do so when test on unseen targets and scenes. Nowadays, researchers have made some efforts to solve this generalization problem and conducted some experiments on embodied AI tasks\cite{oh2016control,parisotto2018neural,zhang2017deep,mousavian2019visual}. In contrast, we carry out our research from the perspective of knowledge graph, because many common scenes (e.g. images or text) can generally be converted into graphs and our visual system also models what we see in the forms of graphs. Therefore, learning from our human navigational mechanism maybe a great way to achieve reasonable navigation performance. For humans, our visual system is so strong that we only take actions that will make us closer to the target location and can navigate freely in unseen environments. We can establish a powerful 3D knowledge base (see Fig. \ref{fig-knowledge-graph}) after exploring numerous environments. The 3D knowledge obtained from our visual system provide us multitudinous information, such as object attributes (e.g. color, shape, size, distance, open, category), obstacle orientation (e.g. table is on the left), local spatial relationships (e.g. left, right, front, under, up, on) between objects in the current filed of view, and global spatial information which summed up for navigation reasoning. For example, from experience, we know that eggs are usually placed in closed fridges while mugs are often next to coffee machine, and hitting a table can cause a fall. Such knowledge summarization ability maybe the essential factor for us to achieve transfer learning, conduct target-induced exploration, and help us avoid obstacles. Therefor, our work intend to incorporate 3D knowledge graph into classical DRL framework to help agents establish such knowledge system for navigation. We extract 3D spatial relationships between objects during agent exploration to form graphs, then apply graph convolutional networks to obtain node features for the established graph. Moreover, our visual system do not actually focus on every part of the entire image when we observe the world, but select some specific parts. For another, we will learn where to focus in the future based on previously observed images. Since our visual system tends to pay more attention to the part information that assists judgment and ignore irrelevant information, we further use attention mechanism that learns to focus on the most important references (objects) in the current filed of view or target image to guide policy search. Our policy ultimately depends on the graph representations together with visual features to make final decisions.

Sparse reward is a classic problem in deep reinforcement learning, and the main reason that makes it difficult to learn. HER \cite{andrychowicz2017hindsight} allows sample-efficient learning from rewards which are sparse and achieves good results on the task of manipulating objects with a robotic arm. Consistent with the principle of HER, we propose a module to solve the sparse reward problem in visual navigation. Our basic idea is that most of the agent's explorations maybe failed, but the agent could still learn from these failed explorations. Inspired by the fact that one trajectory explored by a robot contains not only information from the starting point to the destination, but also information on how to reach the intermediate points of the trip, so if switch targets, then a failed experience can become a successful experience to reach other targets. We introduce a sample augmentation module named target skill extension (TSE) which generates sub-targets during training process and allows learning from failures. From a trajectory explored by the agent, we select the places containing the finding objects or other reasonable objects as new sub-targets, and divide the trajectory into multiple trajectories according to these sub-targets. Finally, our model trains all these divided trajectories to improve sample efficiency and to distill navigation policy. Our experiments show that training sub-targets greatly speed up the training process and has a greater potential to learn more generalized navigation rules.

Instead of considering that navigation terminate automatically when the agent moves sufficiently close to the target, we add a dedicated action into the agent's vocabulary: stop. Task completion notified by the environment do not test whether the agent self-realize that it has reached its destination. To indicate such understanding, our agent needs to issue this stop action when its destination is arrived. Otherwise, navigation will be considered a failure. In addition to focus on static targets that can be found directly in the environment, we also adopt actionable targets to evaluate navigation performance. In real world, what we want robots to look for is not always visible. For instance, eggs are usually placed in a closed fridge and can only be discovered until the agent walks to the front of fridge and opens it. As illustrated in Fig. \ref{fig-target}, we introduce actionable targets that navigation task is not successful until the agent finds its target's receptacle and opens it. What's more, we train our policy also using imitation learning (IL), which utilize a small amount of demonstration data to assist agents. We evaluate our methods in the AI2-THOR \cite{kolve2017ai2}, which provides near photo-realistic indoor environments and support for object manipulation actions. Since currently only the work \cite{zhu2017target} focus on the target-driven visual navigation, we mainly compare results with their work. But we also report the navigation performance using more advanced architectures with LSTM\cite{hochreiter1997long}. The experimental results show that our method achieves significant improvement as compared to the baselines.

\section{Related Work}

\subsection{Visual Navigation}
Traditional approaches decompose navigation task into two sub-tasks by building a 3D map of the scene and then planning in this constructed map \cite{bonin2008visual,gupta2017cognitive,savinov2018semiparametric}. Recent success of deep learning and reinforcement learning has made learning-based navigation approaches more popular. Learning-based navigation tasks can be distinguished along several dimensions, such as visual semantic navigation \cite{zhu2017visual,mousavian2019visual,wu2019bayesian,yang2019visual,wortsman2019learning}, instruction following \cite{mei2016listen,oh2017zero,chaplot2018gated,anderson2018vision,wang2018look,fried2018speaker,wang2019reinforced}, and embodied question answering \cite{gordon2018iqa,das2018embodied}. In our case, the target is given to the agent in terms of image. Compared with instruction following and embodied question answering, we focus on navigation task that evaluation metrics only consider navigation component.

Early learning-based navigation researches aim at navigation in synthetic game-like environments, such as ViZDoom \cite{kempka2016vizdoom}, Minecraft \cite{oh2016control} and DeepMind Lab \cite{beattie2016deepmind}. \cite{kulkarni2016deep} uses a feed forward architecture to learn deep successor representations. \cite{tessler2017deep} incorporates deep skill networks into hierarchical DRL network architecture to achieve knowledge retention. \cite{mirowski2017learning} and \cite{jaderberg2017reinforcement} improve navigation ability in mazes by introducing self-supervised auxiliary tasks. However, these works evaluate navigation performance in same training scenes or same targets. Our goal is to train an embodied AI agent who can transfer learned navigation skills to unseen scenes and targets.

More recent works evaluate the navigation performance in slightly different scenes which not used during training. \cite{oh2016control} examines how well a variety of Q-networks with external memory generalize to unseen or larger maps. \cite{chaplot2016transfer} conducts an empirical study of deep Q-networks to investigate transfer learning. \cite{parisotto2018neural} uses a spatially structured 2D memory image to store information about the environment over long time lags. \cite{oh2017zero} presents a hierarchical architecture where a meta controller learns to use the acquired skills. \cite{chaplot2018gated} uses a gated-attention mechanism to learn a policy. However, these works either focus on over-simplified tasks or test on environments which are only slightly varied from pixel-level variations or small mazes. In contrast, we use more diverse environments which contains different visual and structural observations.

The Latest navigation studies focus on multi-targets tasks and generalization ability with the construction of more realistic  3D simulated environments, such as AI2-THOR \cite{kolve2017ai2}, House 3D\cite{song2017semantic}, Matterport 3D \cite{chang2018matterport3d}. \cite{zhu2017visual} develops a deep predictive model based on successor representations to ensure cross task generalization. \cite{pathak2018zero} learns a goal-conditioned skill policy using data collected by self-supervised exploration. \cite{wortsman2019learning} uses a meta-reinforcement learning approach to learn a self-supervised interaction loss that encourages effective navigation. The research most directly relevant to our work is \cite{zhu2017target}, we both represent our targets as RGB images. They propose an actor-critic model whose policy is a function of the target as well as the current state. However, We use the 3D knowledge graph and sub-targets to boost performance across scenes and targets. What's more, we contain more kinds of actions and include stop action. Especially, we create actionable targets which find some of the targets requires interaction.

\subsection{Deep Reinforcement Learning}
Reinforcement learning is learning how to map situations to actions so as to maximize a numerical reward signal. Deep reinforcement learning methods use deep neural networks to approximate of the following components of reinforcement learning: value function, $V(s;\theta)$ or $Q(s,a;\theta)$, policy $\pi(a|s;\theta)$, and model (state transition function and reward function). Here, the parameters $\theta$ are the weights in deep neural networks that needs to learn. DRL methods have shown success in several domains such as video games \cite{mnih2015human}, chess playing \cite{silver2016mastering}, and continuous control \cite{lillicrap2016continuous}. There are many kinds of deep reinforcement learning algorithms, but the value-based DRL methods are currently the most popular, and we also mainly carry out our experiments using value-based algorithms. \cite{mnih2015human} introduce deep Q-network (DQN) and ignite the field of DRL. DQN use a convolutional neural network (CNN) to approximate the action value function $Q(s,a)$. \cite{van2016deep} propose double DQN (D-DQN) to tackle the over-estimate problem in DQN, they evaluate the policy according to the online network, but to use the target network to estimate value. \cite{schaul2016prioritized} proposed to prioritize experience replay, so that important experience transitions can be replayed more frequently. \cite{wang2016dueling} propose the dueling network architecture to estimate state value function $V(s)$ and the associated advantage function $A(s,a)$. \cite{mnih2016asynchronous} propose asynchronous advantage actor-critic (A3C) to reduce variance and accelerate learning. For Atari games, A3C runs much faster yet performs better than or comparably with DQN, D-DQN, Dueling DQN, and prioritized D-DQN. \cite{schulman2017proximal} propose proximal policy optimization (PPO) to alternate data sampling and optimization by constraining gradient updates. PPO achieves good performance on several continuous tasks in MuJoCo, on continuous 3D humanoid running, steering, and on discrete Atari games. In A3C, parallel actors employ different exploration policies to stabilize training and performs better than the other reinforcement learning techniques because of the diversification of knowledge, we use A3C algorithm as our baseline for our navigation task.

\subsection{Knowledge Graph}
Knowledge graph has become an active research filed in recent years. These graphs may be undirected, directed, and with both discrete and continuous node and edge attributes \cite{niepert2016learning}.  Generally, the knowledge graph contains objects, attributes, and relationships which able to combine multiple levels of scene information in a more coherent way, and hence has quickly gained massive attention. Knowledge graph also has potential to discover hidden knowledge from semantic data structures. At present, considerable works use knowledge graph for computer vision (CV), natural language processing (NLP), and other area tasks. \cite{marino2017more} investigates the use of structured prior knowledge to improves performance on image classification. \cite{wang2018zero} consider the problem of zero-shot recognition, which learns a visual classifier for a category with zero training examples. \cite{norcliffe2018learning} propose a graph-based approach for visual question answering. \cite{hu2018relation} propose an adapted attention module for object detection. \cite{wang2018deep} propose an end-to-end trainable and interpretable graph reasoning model to facilitate social relationship recognition. \cite{gao2019know} propose an end-to-end zero-shot action recognition framework. \cite{yang2019visual} constructs knowledge graph from dataset visual genome \cite{krishna2017visual} as prior knowledge for visual semantic navigation. However, most of these works do not involve the process of edge connection, their knowledge graphs are readily available. In our work, the knowledge graph is represented in the forms of 3D spatial knowledge and gradually self-formed through agent exploration in the environments. Instead of using graph convolustional network to simply extract graph features, we further use the graph representations of the 3D spatial relationships to self-infer important navigation rules, such as, the most relevant references to attend on and avoid obstacles.

\section{The Proposed Method}
\begin{figure*}[!t]
	\centering
	\includegraphics[width=6.5in]{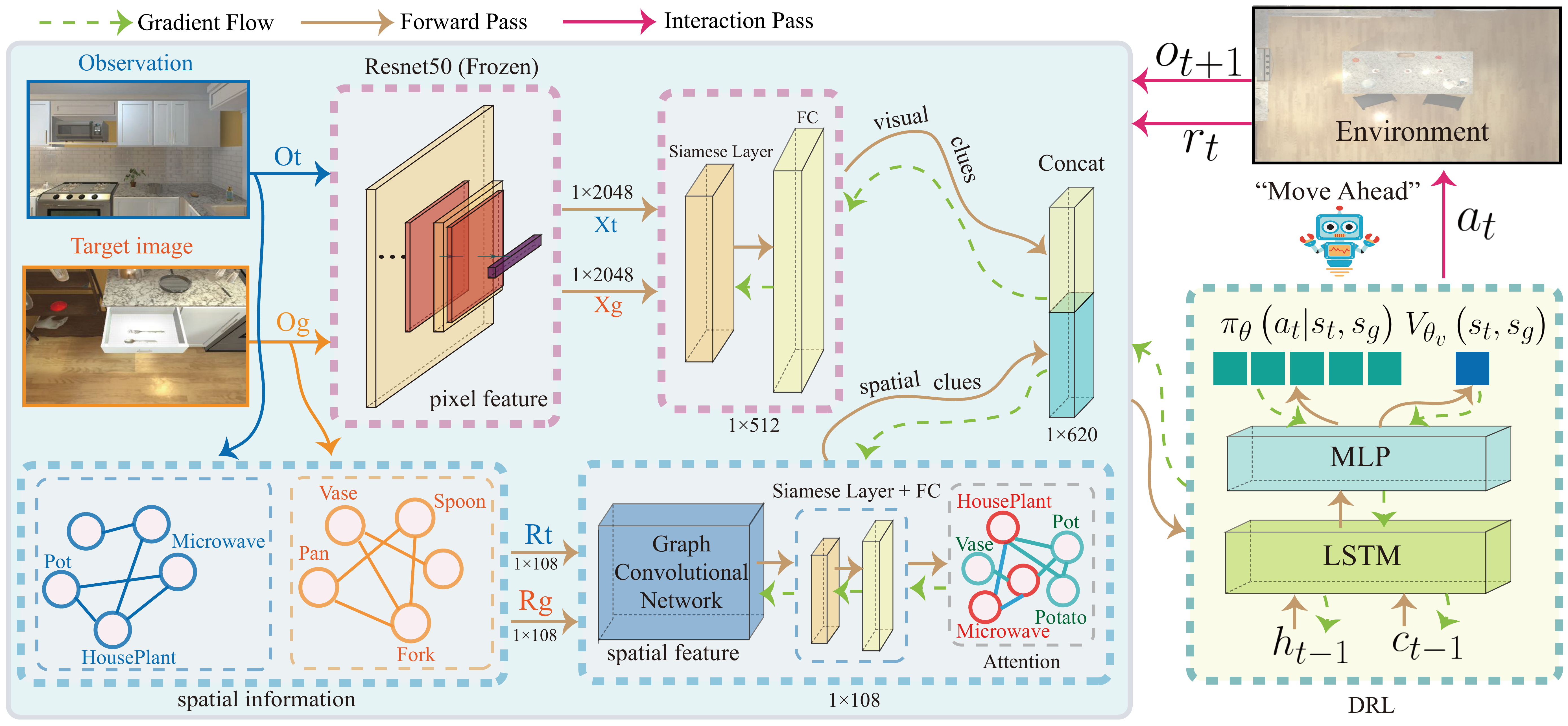}
	\caption{Overview of our proposed navigation framework. Our model incorporate 3D spatial relationships into target-driven visual navigation. The inputs of our network at each time step $t$ are the egocentric images (observation) from the current location and the pre-specified target image. Specifically, we learn a policy $\pi(a_t|s_t,s_g)$ that decides an action based on the visual features together with the spatial features. Visual features are obtained through Resnet50 and then fused by a siamese layer. We use the recently proposed graph convolutional network to extract node features of the graphs from observation and target image. Furthermore, we apply attention mechanism to infer the most important references (objects) to guide decision-making.}
	\label{fig-model}
\end{figure*}

In this section, we present an in-depth description of our proposed method, an architecture that extends an original deep reinforcement learning algorithm, A3C, with the representations of the 3D spatial relationships. The agent constructs its knowledge graph via visual recognition and reasons navigation knowledge through attention mechanism to make decisions. Fig. \ref{fig-model} illustrates the basic framework of our spatial relationships based visual navigation system.

\subsection{Problem Definition}
Our goal is to train an agent which can find the minimum length sequence of actions to reach a pre-specified target location while avoiding obstacles. Our agent perceives its target location and navigating environment both through RGB images. We formulate our navigation task as a partially observable Markov decision process (POMDP) problem in consideration of the sequential-decision making nature of the visual navigation. At each time $t$, the agent receives an observation $o_t$, then produces an action $a_t$. Once the action is taken, the agent receives a scalar reward $r_t$ and a new observation $o_{t+1}$ from the environment. The agent's state $s_t$ is a function of its observation at time $t$, $s_t = f(o_t)$. 

Note that we do not hypothesis the environment will notify the agent when it reaches the target location, since this setting makes the agent does not self-realize whether it has reached the target location or not. In contrast, we consider the stop action and expect the agent to issue this action when it reaches the target location. What ensues is that this makes agent's learning process more challenging due to more sparse reward problem. However, we emphasize that introducing stop action is better for the agent to generalize to real world settings. Our trained agent will really understand what its task is and whether it has finished its task. In addition, our navigation task involves two type of targets: actionable targets and static targets. Static targets refer to the finding objects that can be directly seen through random walking in the environment. Actionable targets are usually hidden in some receptacles (e.g. fridge, drawer, cabinet), and cannot be directly found unless an open operation is taken. Successfully navigating to actionable targets require a deep understanding of its scenes, the agent must learn which objects can be opened and infer which receptacles the finding object may be hidden in.

\subsection{Policy Learning}
We utilize deep reinforcement learning algorithms to learn the optimal navigation policy, which learns a mapping from the 2D image to an action in the 3D scenes. Navigation skills using deep reinforcement learning are learned by maximizing accumulated rewards. Our navigation task can be regarded as a multi-targets learning problem, so our policy needs to condition on both the input and the target. Our policy takes the current observation $o_t$ and the finding target image $o_g$ as input to make decisions. The agent's action $a_t$ at each time step $t$ is determined by a parametrized policy function $\pi(a_t|s_t,s_g;\theta)$ which allow our trained agent to generalize across multiple targets. We adopt the asynchronous advantage actor-critic (A3C) \cite{mnih2016asynchronous} algorithm that relies on learning both a policy $\pi(a_t|s_t,s_g;\theta)$ and value function $V(s_t,s_g;\theta_{v})$ given the current state $s_t$ and target state $s_g$. A3C optimizes the policy by minimizing the loss function

\begin{equation}
L_{\pi}(\theta)=-\mathbb{E}[\sum_{t=1}^{T}(R_t-V(s_t,s_g;\theta_{v}))log\pi(a_t|s_t,s_g;\theta)]
\end{equation}

where $\theta$ and $\theta_{v}$ are the parameters of actor network and critic network respectively. $T$ is the length of the explored trajectory. $R_t$ represents $k$-step return and is the discounted accumulative reward defined by

\begin{equation}
R_t=\sum_{i=0}^{k-1}\gamma^{i}r_{t+i}+\gamma^{k}V(s_{t+k},s_g;\theta_{v})
\end{equation}

where $\gamma \in (0,1)$ is the discounted factor that reflects the significance of future rewards. The value function is updated by minimizing the loss

\begin{equation}
L_v(\theta_{v})=\mathbb{E}[(R_t-V(s_t,s_g;\theta_{v}))^2]
\end{equation}

Finally the overall loss function for A3C is $L_{A3C}(\theta, \theta_{v})=L_\pi(\theta)+\alpha L_v(\theta_{v})$, where $\alpha$ is a constant coefficient.

In A3C many instances of the agent interact in parallel with many environments, which both accelerates and stabilizes learning. The A3C architecture we build on uses an LSTM \cite{hochreiter1997long} and two MLPs to jointly approximate both policy $\pi(a_t|s_t,s_g;\theta)$ and value function $V(s_t,s_g;\theta_{v})$.

\subsection{Visual Features}
Deep convolutional neural network (CNN) today has shown excellent performance in multiple computer vision tasks, such as image classification, object localization, and instance segmentation. He et al. \cite{he2016deep} presented a residual learning network (ResNet) to learn features from visual images and won the first place in the ILSVRC 2015 image classification task. We use ResNet50 which trained in the ImageNet\cite{deng2009imagenet} database to extract visual features $X_t$ for each input observation $o_t$. Besides, we apply same module to derive representation for the target image. Afterwards, we concatenate visual features from the current observation and the target image using a deep siamese network \cite{chopra2005learning}. Siamese framework is a method for training a similarity metric from data. Similar to \cite{zhu2017target}, the image representations of the current observation and the target are transformed into the same embedding space using the two streams of weight-shared siamese layer. Then information from both embeddings is fused through one fully-connected neural network to form a joint representation $I_t$. Our final visual features at time step $t$ is obtained by

\begin{gather}
X_t=f(o_t;\theta_{ResNet}) \\
X_g=f(o_g;\theta_{ResNet}) \\
I_t=\sigma(W_f[\sigma(W_sX_t),\sigma(W_sX_g)])	
\end{gather}

where $\theta_{ResNet}$ is the parameters of ResNet50 network, $W_s$ is the parameters of the siamese network and $W_f$ is the parameters of the fusion network. $\sigma$ denote the ReLU activation function and [ , ] represents concatenation.

\subsection{Spatial Relationships}
Most current researches learned their navigation policy directly based on the pixel-level features, and usually struggle against poor generalization ability when test on unseen targets and scenes\cite{dhiman2018critical}. In addition to extract visual features, we propose a model that also incorporates 3D knowledge graph into original deep reinforcement learning frameworks. We use 3D knowledge graph potentially to assist the agent avoid collisions and guide the agent to reach the target location in a shortest path way. Moreover, we use recently proposed graph convolutional networks (GCN) \cite{kipf2017semi} to compute spatial representations for decision making.

\begin{figure}[!t]
	\centering
	\includegraphics[width=3.5in]{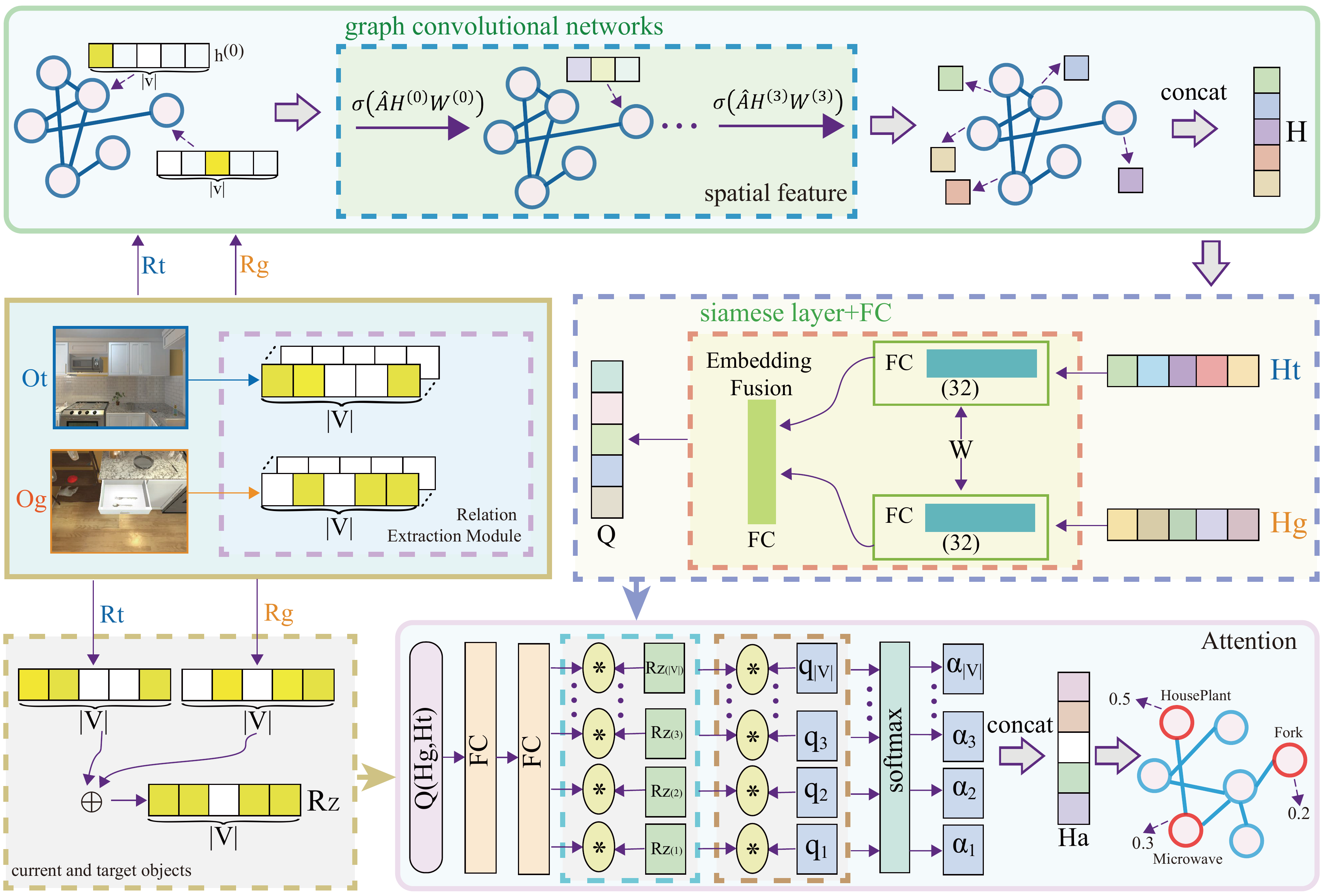}
	\caption{Illustration of our 3D spatial representations. We build 3D knowledge graph from the relation extraction module while agent exploring its environments. Then, the recently proposed GCN is used to encode spatial relationships between different objects. In addition, the final spatial representations is obtained based on the spatial features from the current observation and the target image. Finally, we use attention mechanism to self-infer the most relative references (objects) from the final spatial representations for avoiding obstacles or guiding policy search. The attended feature vector $H_a$ and the visual features are cascaded to produce the next action.}
	\label{fig-graph}
\end{figure}

Note that 3D knowledge graph could encode multiple spatial relationships (e.g. front, left, right, in, up, under) between objects as shown in Fig. \ref{fig-knowledge-graph}, and our spatial knowledge for navigation needs to be learned during agent exploration. We denote 3D knowledge graph by $G=(V,E)$. Specifically, each node $v\in V$ denotes an object, and each edge $e \in E$ denotes a type of relationship between a pair of objects. Our agent needs to be equipped with two important skills to construct our 3D knowledge graph. One is that our agent must know what objects are in the current observation and the target image. Besides, our agent could also learn what are the relationships between these objects. Because object detection is not our main task, and \cite{gordon2018iqa} have demonstrated that the YOLOv3 \cite{redmon2018yolov3} fine-tuned on the training scenes in the AI2-THOR could be a good object detector, we directly obtain object information from the AI2-THOR to solve the first problem. Remarkably, the object information provided by the environment still requires our agent to learn common sense. For instance, our agent must learn that microwaves can be opened, mugs often next to a coffee machine, moving forward is not allowed cause a table in front. To solve the second problem, we only consider the adjacency relationships between objects in our current work. Our 3D knowledge graph include all objects that appear in the AI2-THOR and two nodes are connected with an edge if they are both visible. An object is considered visible if its distance is within 1.5 meters from the agent's camera. Then our 3D knowledge graph is presented in the form of adjacency matrix. In detail, as shown in Fig. \ref{fig-graph}, the relation extraction module firstly extract multiple relationships between visible objects. Specifically, we denote visible object information by a binary vector $R$, the binary value of each position in $R$ indicates whether the corresponding object is visible. We denote the total visible objects $R_z$ from visible objects  $R_t$ in the current observation and visible objects $R_g$ in the target image. After that, we build the adjacency matrix $A$ based on the constructed knowledge graph and initialize each node feature $H^{(0)}$ as a one hot feature vector. $\hat{A}$ is obtained by performing normalization on $A$ to make each node contain its own node features. The core idea behinds graph neural networks is to use edge information to aggregate node information to generate new node representations. We further perform spatial information propagation to compute the node feature vectors $H$ using GCN \cite{kipf2017semi}:  

\begin{equation}
H^{(l+1)}=\sigma(\hat{A}H^{(l)}W^{(l)})
\end{equation}

where $H^{(0)}=I$, $I$ is an identity matrix, and $W^{(l)}$ is the parameter for the $l$-th GCN layer. $\sigma$ denote the ReLU activation function. The final obtained node features $H$ encode the spatial knowledge for the navigation task.

Since our visual system pay more attention to the part information that assists judgment and ignore irrelevant information, we adopt attention mechanism to summarize spatial knowledge for navigation. For instance, if the target is often next to shelf and shelf is on the left, there are no obstacles on the left as well, then the next action should move left. Some times, maybe a few key objects in the image can make a deterministic decision. Seen in Fig. \ref{fig-graph}, let $H_t$ and $H_g$ encode the spatial features from GCN for the current observation and the target image respectively. Let final attention vector $H_a$ be the dynamic probability representations of the most relevant objects of the images at time step $t$ to find the target location. For each object $i$, the attention mechanism module generates a positive weight $\alpha_i$ which can be interpreted the probability that object $i$ is the right object to focus on for producing the next action. Because agent's next action is determined by the target and the current observation, the weight $\alpha_i$ of each object is conditioned on spatial features $H_t$ and $H_g$. Just like visual features, we use the same fusion ways to get the fusion spatial features $Q$. We constraint the choice of attention to objects in the current observation and target image. Our agent also learns what is the most appropriate references among target objects. Then we use two fully connected (FC) layers to compute final spatial representations $H_{a(i)}$, $i=1,..,|V|$ corresponding to the probability extracted for different objects. The final attention vector $H_a$ could be described as:

\begin{gather}
Q=W'_f[\sigma(W_{s}'H_t),\sigma(W_{s}'H_g)] \\
Q'=(QW_{fc1}W_{fc2}R_z)Q\\
H_{a(i)}=\frac{\mbox{exp}(q'_i)}{\sum_{i=1}^{|V|}\mbox{exp}(q'_i)}
\end{gather}

where $W_s'$ and $W_f'$ are the parameters of siamese and fusion networks respectively. $W_{fc1}$ and $W_{fc2}$ are the parematers of FC layers. $\sigma$ denote the ReLU activation function. $Q=[q_1,...,q_{|V|}]$ is the fusion spatial vector.

Finally, the joint representation of the visual features and the spatial representations are fed into the DRL module to obtain an action output ${a_t}$.

\subsection{Sub-targets Extraction}
\begin{figure}[!t]
	\centering
	\includegraphics[width=3.5in]{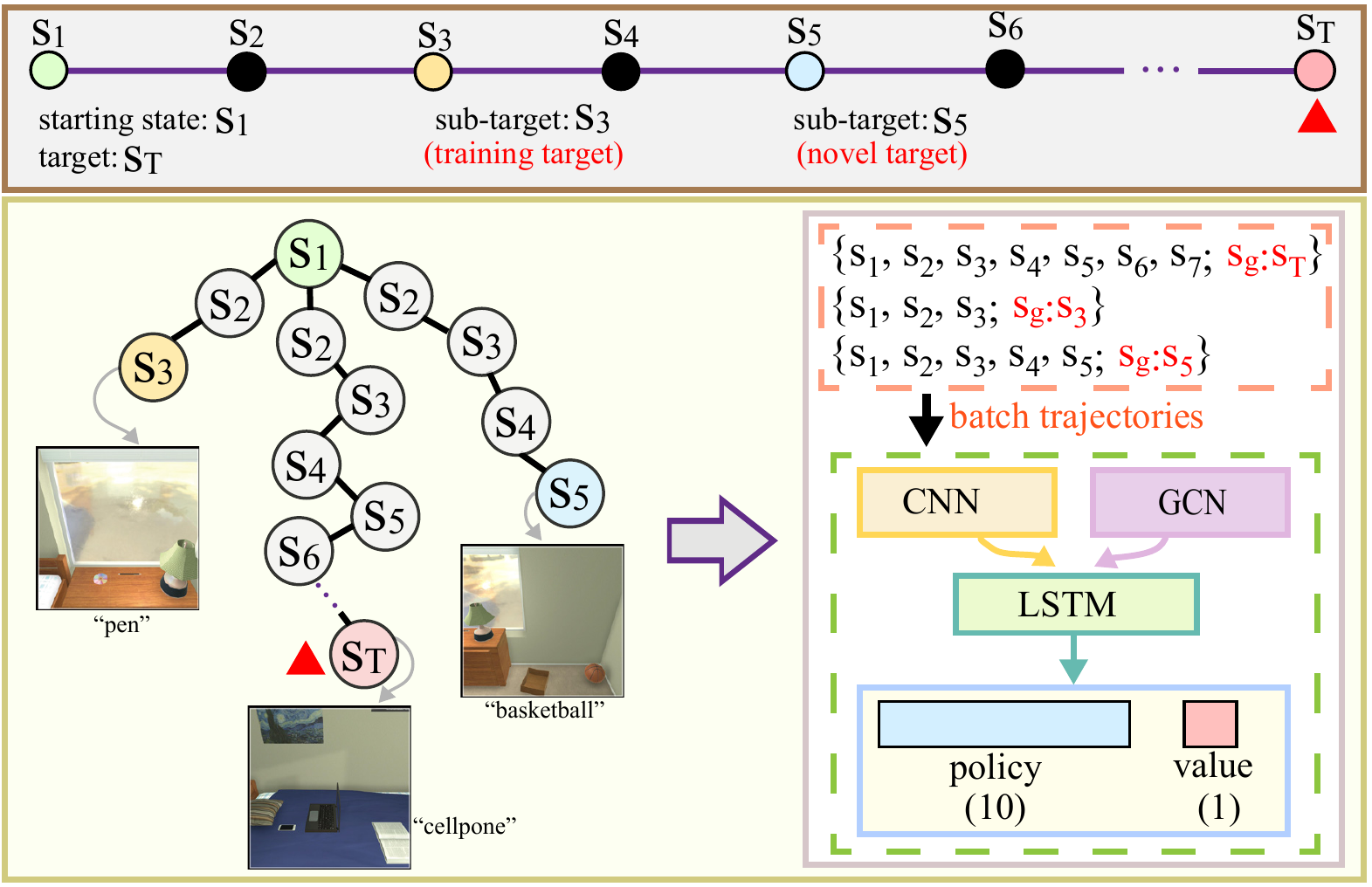}
	\caption{Our TSE module decomposes an explored trajectory into some reasonable trajectories by selecting encountered sub-targets where they can be the targets (e.g. $s_3$) trained by other threads or other reasonable targets (e.g. $s_5$) which not used for training. Finally, our A3C model trains all divided trajectories to speed up training process and improve data efficiency.}
	\label{fig-subtarget}
\end{figure}

Sparse reward is a classic problem in deep reinforcement learning,  agent can not be rewarded frequently due to the huge searching space.
Inspired by the fact that one trajectory explored by a robot contains not only information from the starting point to the destination, but also information on how to reach the intermediate points of the trip, so if switch targets, then a failed experience can become a successful experience to reach other targets. This way our agent could also learn from its failures. Besides, in our settings, the A3C algorithm learns a policy with multiple threads and each thread learns for a different target. However, the target learned by one thread may still be encountered by other threads. We extend traditional A3C to incorporate sub-targets for data augmentation and term this the target skill extension (TSE) module. Suppose $<s_1, a_1, r_1, ..., s_{T}, a_{T}, r_{T}>$ is a trajectory obtained by the agent after it explored the environment, where $s_1$ is a random start state and $s_T$ is a terminal state. Previous approaches only acquire the knowledge of how to go from $s_1$ to $s_T$, because they only train the policy $\pi(a_t|s_i,s_T;\theta)$ ($i\leq T$), and then discard the trajectory and start the next new round of exploration. Nevertheless, such training way does not make use of the explored experience. For each thread, in addition to train $\pi(a_t|s_i,s_T;\theta)$, our TSE module also train the policy $\pi(a_t|s_i,s_j;\theta)$ ($i\leq j$) if $s_j$ is a reasonable sub-target. There are many possible ways to define sub-targets and then divide a trajectory into multiple trajectories according to them. For each sampled trajectory, as illustrated in Fig. \ref{fig-subtarget}, we select some observations contained at least one novel object (e.g. $s_5$) or the targets trained by other threads (e.g. $s_3$) as sub-targets. Novel objects refers to the objects that can be picked up and is not used for training. In this way, we divide a trajectory into multiple trajectories of different targets for training. Every time after we train a sub-trajectory, we always assign the parameters of global network to local networks.

\subsection{Imitation Learning}
Since training an agent from scratch is usually infeasible using pure deep reinforcement learning approaches due to huge state space and very sparse rewards, we also use imitation learning to speed up training. We utilize only one expert trajectory for each training target to assist agents. Generating shortest path is usually not easy and there are multiple reasonable ways of going from one state to another. In contrast, we generate expert trajectories from the target location $s_T$ to a random starting point $s_1$ by limit action space (move back, move left, rotate right) and their priorities. We select one action that the coordinate of the environment will change after taking this selected action $a_t$. Then we reverse the trajectory to get a trajectory of length $N$ from the starting point $s_1$ to the target $s_T$. Our path generation method can ensure us to get an approximate optimal trajectory. Instead of employing a two-stage training process, we train the expert trajectories in a online fashion and before a certain number of time steps.

\section{Experiment}
In this section, we present the navigation results in comparison to several baselines without 3D knowledge graph and sub-targets. We evaluate our method by testing its generalization performance against unseen targets and unseen scenes. Our results provided by our experiments show that 3D knowledge graph and sub-targets is useful.

\subsection{Experiment Setup}
We evaluate our methods in the 3D indoor environment, AI2-THOR \cite{kolve2017ai2}, which covering four different room categories: kitchens, living rooms, bedrooms, and bathrooms. Each room has a set of objects and rich styles. Compared with other 3D environments, AI2-THOR allows the agent to perform several actions to interact with its scenes, such as open, pick up, and push. Some objects in AI2-THOR are not directly visible if there are no interactions. For instance, cups are not visible since they always hidden in closed cabinets at the beginning of random scene initialization. In our navigation settings, the agent can perform two interaction actions (open, close) to manipulate objects. A wide variety of objects, such as fridges, cabinets, and drawers, can be interacted with. There are a total 108 different objects in the AI2-THOR, so our 3D knowledge graph have $|V|=108$ objects.

Our navigation task is to train an agent that can navigate from a random starting position to a pre-specified target location only via egocentric view. Specially, our agent perceives its navigation target through a pre-specified target image. We select the objects that can be picked up (e.g. apple) or belong to common household items (e.g. microwave) as navigation targets. We select one nearest location as the destination for our static targets and actionable targets. Our actionable targets are likely to hide in receptacles which our agent can not see them directly. In order to successfully find actionable targets, our agent must learn affordances, for example, fridges can be opened, and apples are often placed in fridges.

Our navigation task is considered successful if the agent go to the destination and perform the stop action meanwhile. We use navigation, view and interaction commands of the AI2-THOR to conduct our experiments. We consider the action spaces $A=\{$move forward, move back, move right, move left, rotate right, rotate left, look up, look down, open object and stop$\}$. We do not include close action because our agent automatically apply close operation if it does not plan to apply stop action at next time when it has opened a receptacle. We provide a target-reaching reward (10.0) upon task completion and a terminal reward (0.01) when our agent arrive at the destination. To encourage shorter trajectories, we add a time penalty (-0.01) as immediate reward.

\begin{table*}[!t]
	\renewcommand{\arraystretch}{1.3}
	\caption{SPL/SR (\%) results are shown for our model and baselines. We compare against a random walk baseline, imitation learning, FF A3C\cite{zhu2017target} and other LSTM models. All the accuracy values are averaged over 100 episodes for each target.}
	\label{tab-result1}
	\centering
	\begin{tabular}{c||c|c|c||c|c}
		\hline
		&
		\multicolumn{3}{c||}{static targets}&
		\multicolumn{2}{c}{actionable taregets}\\
		\hline
		&
		seen scenes,&
		seen scenes,&
		unseen&
		seen scenes,&
		unseen\\
		&
		seen targets&
		unseen targets&
		scenes&
		seen targets&
		scenes\\
		\hline
		Random walk&
		0.11/1.56&
		0.26/14.86& 
		0.29/15.84& 
		0.02/0.23& 
		0.08/3.25\\
		\hline
		Imitaiton learning&
		2.22/8.24&
		0.72/5.96& 
		0.77/4.09& 
		1.04/5.54& 
		0.01/0.25\\
		\hline
		LSTM A3C&
		0.08/0.2& 
		0.12/0.29& 
		0.14/0.31& 
		0.0/0.0& 
		0.0/0.0\\
		\hline
		FF A3C&
		35.16/88.33& 
		10.81/35.64& 
		--/--& 
		19.92/65.13& 
		--/--\\
		\hline
		LSTM A3C+IL& 
		46.04/93.19& 
		11.72/36.58& 
		5.15/18.03& 
		32.82/84.6&
		1.94/13.75\\
		\hline
		LSTM A3C+TSE& 
		42.58/93.16&
		10.82/35.0&
		4.31/16.69&
		32.51/84.3&
		1.35/13.37\\
		\hline
		LSTM A3C+IL+TSE&
		49.74/98.41& 
		12.77/37.64& 
		5.21/20.97& 
		36.06/86.6& 
		2.41/15.13\\
		\hline
		LSTM+KG&
		52.32/98.47&
		14.1/41.95&
		6.11/32.56&
		36.77/87.55&
		3.23/17.25\\
		\hline
		LSTM A3C+KG+Attention&
		\textbf{52.58/98.44}&
		\textbf{14.89/44.25}& 
		\textbf{7.20/41.09}& 
		\textbf{38.51/88.78}& 
		\textbf{5.26/21.13}\\
		\hline
	\end{tabular}
\end{table*}

Because kitchen rooms contain more actionable targets than other type of rooms in the AI2-THOR, we evaluate our methods in all kitchen rooms. We split the kitchen rooms into three splits, 20 training rooms, 5 validation rooms and 5 testing rooms. Since there are approximately 2 actionable targets for each kitchen room, we choose 20 rooms with more actionable targets as training rooms. All actionable targets in the training scenes are used for training. In practice, our TSE module randomly sample 5 sub-targets form each explored trajectory. All sub-target trajectories along with the explored trajectory are used for training.

\subsection{Evaluation Metrics}
We evaluate our method based on two metrics: Success Rate (SR) and the Success weighted by Path Length (SPL). SR is defined as $\frac{1}{N}\sum_{i=1}^{N}S_{i}$, which is the ratio of the number of times the agent successfully navigates to the target and the total number of episodes. $N$ is the number of episodes and $S_{i}$ is a binary indicator of success in episode $i$. SPL is defined as $\frac{1}{N}\sum_{i=1}^{N}S_{i}\frac{L_{i}}{\max(P_{i},L_{i})}$, which is recently proposed by \cite{anderson2018evaluation} and considers both success rate and the optimal path length. $P_{i}$ represents the length of the path actually taken by the agent and $L_{i}$ is the shortest path distance from the agent's starting position to the target in episode $i$.

\subsection{Baselines}
We compare the navigation performance to the following models: (1) \textbf{Random walk}, the agent randomly picks an action from the action spaces at each time step. (2) \textbf{Imitation learning}, its policy is only trained with imitation learning (IL). We use the demonstration actions provided by our path generation algorithm to conduct supervised learning with the cross entropy loss. (3) \textbf{FF A3C} \cite{zhu2017target}, which uses feed-forward (FF) networks that take as input concatenated features from last-n frames and target image to predict the next action. Because their model consists of the scene-specific layers, which makes their model lack of generalization ability to unseen scenes. (4) \textbf{LSTM A3C}, which uses the same siamese fusion methods in \cite{zhu2017target} to get the visual features. Then the visual features are passed through a shared LSTM to predict the next action. (5) \textbf{LSTM A3C+IL}, which incorporate imitation learning into LSTM A3C. This model is trained not only using online imitation learning, but also using A3C algorithm. (6) \textbf{LSTM A3C+TSE}, which incorporate the TSE module into LSTM A3C. This model is also trained with sub-targets that extracted from explored trajectories. (7) \textbf{LSTM A3C+IL+TSE}, which incorporate both imitation learning and the TSE module into LSTM A3C. (8) \textbf{LSTM A3C+KG}, which incorporate 3D knowledge graph (KG) into LSTM A3C+IL+TSE. It learns the navigation policy according to visual features and the fusion spatial features $Q$. (9) \textbf{LSTM A3C+KG+Attention}, which is our final proposed model. It adds the attention mechanism into LSTM A3C+KG and produce actions based on the final spatial representations and visual features.

\subsection{Results}
For all learning models, we report their performance after being trained with 40M frames (across with all threads). We train our model with 140 threads, each thread learns for a different target. All episodes have a maximum length of 5000 time steps for each training thread. We implement our models in Tensorflow \cite{abadi2016tensorflow} and train them on an Nvidia GeForce GTX Titan RTX GPU. For evaluation, we select the model that performs best on the validation set in terms of success and run 100 different episodes for each target. To be fair, the initial locations of the agent is randomly chosen and all models are evaluated using the same set. Besides, the initial location is at least 10 steps away from the target location. An testing episode ends when either the agent reaches the target location, or after it takes the maximum number of steps. The maximum steps is set to 1,00 for seen targets within seen scenes, and to 1,000 for unseen targets within seen and unseen scenes. Since FF A3C \cite{zhu2017target} uses different policy networks for different scenes, their model lacks of the generalization ability to unseen scenes unless fine-tuned. Our model and other baselines use a single policy network for different scene examples, which more compact and generalizable.

Table \ref{tab-result1} summarizes the results of our proposed model and the baselines. We show the performance on seen targets within seen scenes, next their generalization capabilities on unseen targets within seen scenes. Finally, we investigate the generalization ability to transfer the learned navigation skills trained on a set of training scenes to previously unseen scenes. In our setup, our targets are divided into two types, static targets and actionable targets. Static targets can be directly visible without any special operations, while actionable targets are often hidden in closed receptacles and require an open operation to find them. We provide the performance of static targets and actionable targets separately, but we only train one model for these two type of targets over all scenes.

As shown in Table \ref{tab-result1}, the performance is very poor when agent apply random walk or trained only by imitation learning. Due to huge searching space, random walk achieved very low probability for agent to reach the target. Imitation learning, which trained only with a small amount of expert data, also lacks of navigation capability. We also show the comparison results between our model and pure A3C methods. We can see that a direct application of LSTM A3C does not yield sensible performance, the agent exhibit no learning even after millions of training frames. Moreover, FF A3C using scene-specific layer can learn a good policy compared with LSTM A3C, but is slower due to larger learning parameters. FF A3C seems to not converge after 40M training frames. 

The model trained by LSTM A3C+IL and LSTM A3C+TSE performs better than FF A3C, this indicate that expert data and sub-targets have a significant impact on accelerating the agent's learning rates. Dealing with sparse rewards is one of the biggest challenges in reinforcement learning. To some extent, both expert data and sub-targets are effective in solving the sparse rewards problem. LSTM A3C+IL+TSE that combines IL and TSE module together achieves slightly higher performance than only use one of them. The model, LSTM+KG, which also trained with spatial representations compared with LSTM A3C+IL+TSE, acquires significantly better navigation performance. Furthermore, our final model, LSTM A3C+KG+Attention, achieves best results which indicates the effectiveness of the attention mechanism.

We observe that all models obtain relatively high performance when test on seen scenes and seen targets than across unseen targets. In addition, the scenarios in which both scenes and targets are unseen is more challenging, the performance degrades drastically for both baselines and our proposed models. These results indicate that all methods are prone to over-fitting and trained agent does not really understand its environments. However, we observed that our model improved the performance on the unseen scenes by over 20\% (static targets) and 6\% (actionable targets) as compared to models trained without the 3D knowledge graph. We assume that this is because the 3D knowledge graph facilitate learning generalizable knowledge for navigation. The results demonstrate that the 3D knowledge graph can extract information that is critical for policy learning from the perceptible environment.

Compare static and actionable targets, the performance of the actionable targets is significantly lower than the static targets for all methods. There perhaps maybe two problems for this. One is that the agent needs to infer which receptacles the finding target located in. Another is that there are many receptacles of the same type in the environment, such as many drawers and cabinets, the agent needs to search one by one and remember which ones it has been searched before.

Note that SPL is a rather stringent measure. Among all the methods, We observe that the agent achieves very bad performance when evaluate with SPL metric, which shows that it is very difficult for agent to find the targets in the shortest way.  At best, an SPL of 50\% is expected  to be a good level of navigation performance \cite{anderson2018evaluation}. In our case, field of view for the agent’s camera is orthographic, so there is no overlap between two consecutive observations when taking a rotation operation. Thus, contrary to the shortest path which may achieved by our humans, the agent needs more exploration time steps to integrate information to make the series of decisions.

\begin{figure}[!t]
	\centering
	\includegraphics[width=3.5in]{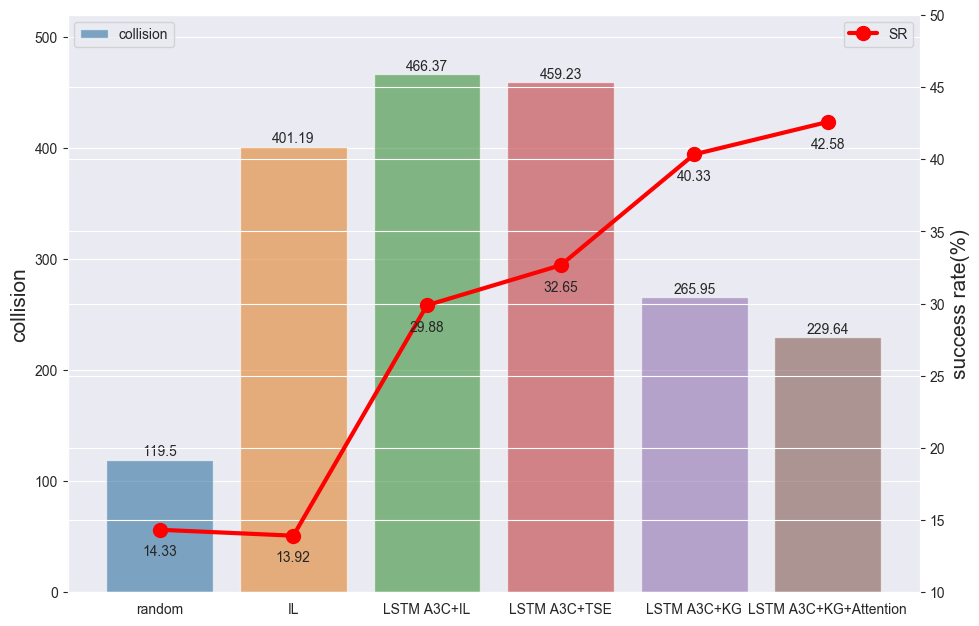}
	\caption{Comparison of the average number of collisions at a trajectory (Bar chart) and the success rate of the target object which can be seen after taken the last action in unseen scenes (Red line). Fewer collisions and higher success rate indicates better navigation performance. Our KG models achieve better performance than the comparative baselines.}
	\label{fig-collision-sr}
\end{figure}

We also show the average number of collisions and the success rate of the last frame which contained at least one target object at the end of trajectory in unseen scenes. From the results in tabel \ref{tab-result1}, as the LSTM A3C seems no learning and FF A3C lacks of generalization ability in unseen scenes unless fine-tuned, their results are not given in current situations when considering performance in unseen scenarios. As shown in Fig. \ref{fig-collision-sr}, our model get better results compared with other listed four baselines. This indicate that the navigation system established by the 3D knowledge graph helps to reduce collisions and conduct target-induced explorations.

\begin{figure*}[!t]
	\centering
	\includegraphics[width=6.5in]{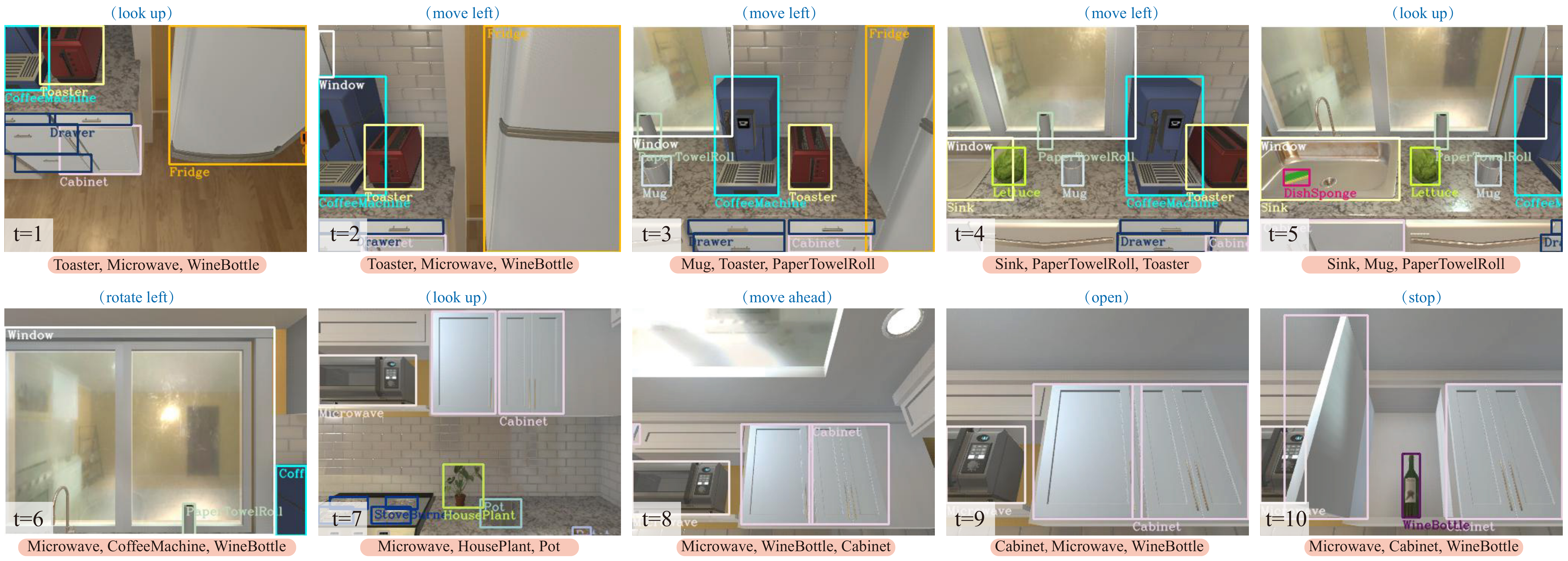}
	\caption{Example results of the attended objects predicted by our model at each time step $t$ and the corresponding action to be taken at the next time. The trained agent self-infers what objects it should attend on from the representations of the 3D knowledge graph to produce the next action. We visualize top-3 objects with the highest probability that the agent inferred at each time step. Our results show that the predicted objects indeed help the agent explore the environment, avoid obstacles, and guide policy search.}
	\label{fig-relation}
\end{figure*}

As illustrated in Fig. \ref{fig-relation}, We show an example of our model's spatial knowledge prediction in a kitchen room. At each time step $t$, our agent select some most relative objects for producing its next action. Note that these attended objects could come from the current observations for reasoning or obstacle avoidance, or could also come from the target image. Since our target image can also contain multiple objects, the agent is also needs to select the most informative objects from that. What's more, this way also  gives a clear indication of the ultimate targets it is looking for. We observe that our agent could properly attend on appropriate objects which lead to next action just like humans. For example, at time step $1$, our agent attend most on the toaster when plan to take the look up action. At time step $3$, our agent attend on the mug when plan to take the move left action. Besides, we notice that agents tend to remind themselves of what they are looking for almostly at every time step. Note that our attention model is end to end trainable and we do not apply any additional attention loss functions. It is totally self-inferred according to the interactive loss. Our results indicate that self-constructed spatial relationships can form general reasoning and has potential to transfer learned skills when more complex 3D knowledge graphs are used in the future.

\begin{figure}[!t]
	\centering
	\includegraphics[width=3.5in]{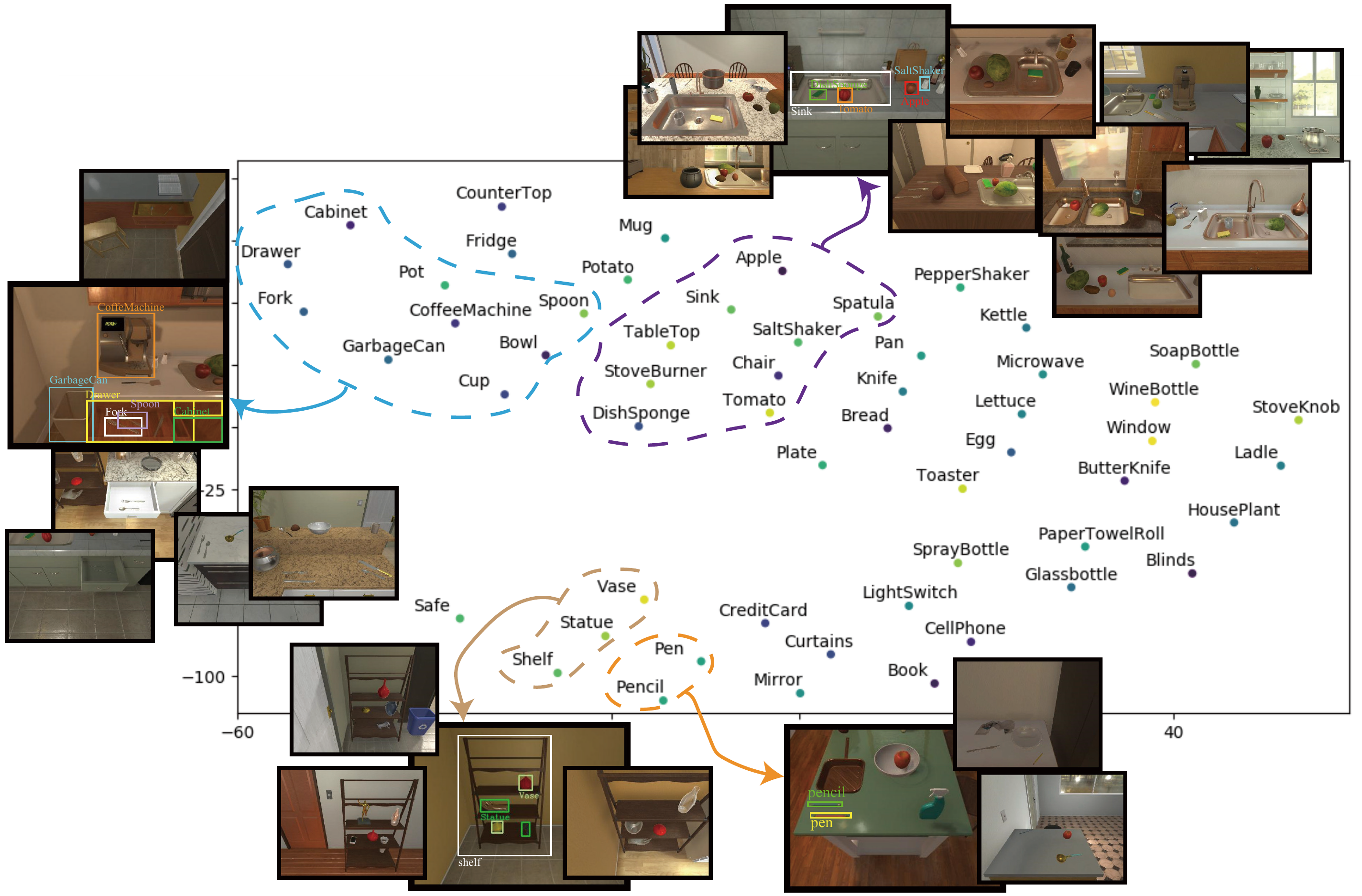}
	\caption{t-SNE embedding of the total objects appeared in all kitchen rooms. We show the node features extracted by the third GCN layer and project them into 2D space. The result indicates that our model has learned relative spatial layout afer exploring numerous rooms.}
	\label{fig-objects}
\end{figure}

In order to understand what our spatial model learns, we examine the node feature vectors learned by our GCN layers. Fig. \ref{fig-objects} shows the t-SNE \cite{maaten2008visualizing} visualization of the node feature vectors obtained from the third GCN layer. We observe that the spatial arrangement of these node feature vectors is commonly consistent with their corresponding 2D projections from t-SNE. For example, vase and statue are usually placed on shelf, so from the picture we can see that the distances between their feature spaces are also close together. This means that our model have learned to project objects into feature space while retaining their spatial configuration.

\begin{figure}[!t]
	\centering
	\includegraphics[width=3.5in]{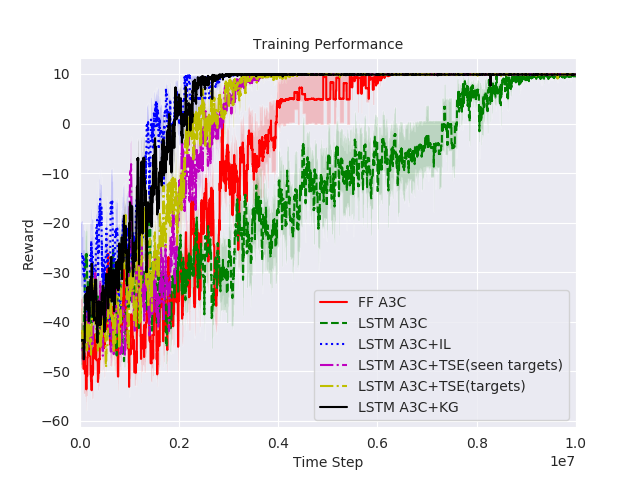}
	\caption{Training curves for our proposed TSE module and baselines. LSTM A3C with sub-targets and imitation learning both converge faster compared to two baselines with standard A3C after 10M training frames.}
	\label{fig-curve}
\end{figure}

Finally, we demonstrate the impact of sub-targets and imitation learning on data efficiency. Our experiments were conducted with 10M training frames and 13 targets (7 static targets and 6 actionable targets) in a kitchen room. As shown in the training curves in Fig. \ref{fig-curve}, FF A3C trained with scene-specific layers converges faster than LSTM A3C using totally generic layers. LSTM A3C with imitation learning greatly speed up the learning process, even though the expert trajectories generated by our path algorithm are suboptimal paths. Besides, LSTM A3C with TSE module also converges more quickly than FF A3C and LSTM A3C, which indicates that sub-targets could be a good alternative to expert data in navigation tasks, at least in terms of data efficiency. We also test whether different types of sub targets influence the training results. TSE(seen targets) refers to LSTM A3C trained with sub-targets that only include training targets. TSE(targets) refers to LSTM A3C trained with sub-targets that include both training and unseen targets. Both of these two different types of sub-targets achieve great convergence results. The Experimental results prove that our TSE module really make full use of the experienced data by learning from failures, and can solve the problem of reward sparsity in traditional reinforcement learning. We also shows the results of LSTM A3C+KG which augment with spatial information and converges normally as well.

\section{Conclusion}
In this paper, we propose an efficient model to solve the problem of target-driven visual navigation, an embodied AI task where an agent must intelligently navigate to a destination through first-person vision according to an assigned target image. In our approaches, we integrate 3D knowledge graph and sub-targets into classic deep reinforcement learning framework to boost navigation performance. Sub-targets are generated by our TSE module and allow agent to learn from failures. Specifically, we use graph convolutional networks and attention mechanism both to form spatial knowledge reasoning and guide policy search. Furthermore, we consider stop action and actionable targets when evaluating navigation performance. Our experiments, which evaluated in the AI2-THOR, show that self-inferred spatial knowledge improves generalization ability across targets and scenes. Our experiments also validates sub-targets can greatly address the data efficiency problem. In our future research, we will try to conduct experiments with more complex types of spatial relationships for better results.

\ifCLASSOPTIONcaptionsoff
  \newpage
\fi



\bibliographystyle{IEEEtran}
\bibliography{IEEEabrv,./reference.bib}

\begin{thebibliography}{10}
\providecommand{\url}[1]{#1}
\csname url@samestyle\endcsname
\providecommand{\newblock}{\relax}
\providecommand{\bibinfo}[2]{#2}
\providecommand{\BIBentrySTDinterwordspacing}{\spaceskip=0pt\relax}
\providecommand{\BIBentryALTinterwordstretchfactor}{4}
\providecommand{\BIBentryALTinterwordspacing}{\spaceskip=\fontdimen2\font plus
\BIBentryALTinterwordstretchfactor\fontdimen3\font minus
  \fontdimen4\font\relax}
\providecommand{\BIBforeignlanguage}[2]{{%
\expandafter\ifx\csname l@#1\endcsname\relax
\typeout{** WARNING: IEEEtran.bst: No hyphenation pattern has been}%
\typeout{** loaded for the language `#1'. Using the pattern for}%
\typeout{** the default language instead.}%
\else
\language=\csname l@#1\endcsname
\fi
#2}}
\providecommand{\BIBdecl}{\relax}
\BIBdecl

\bibitem{zhu2017target}
Y.~Zhu, R.~Mottaghi, E.~Kolve, J.~J. Lim, A.~Gupta, L.~Fei-Fei, and A.~Farhadi,
  ``Target-driven visual navigation in indoor scenes using deep reinforcement
  learning,'' in \emph{Proc. IEEE Int. Conf. Rob. Autom. (ICRA)}, May. 2017,
  pp. 3357--3364.

\bibitem{anderson2018vision}
P.~Anderson, Q.~Wu, D.~Teney, J.~Bruce, M.~Johnson, N.~S{\"u}nderhauf, I.~Reid,
  S.~Gould, and A.~van~den Hengel, ``Vision-and-language navigation:
  Interpreting visually-grounded navigation instructions in real
  environments,'' in \emph{Proc. IEEE Conf. Comput. Vis. Pattern Recognit.
  (CVPR)}, Jun. 2018, pp. 3674--3683.

\bibitem{das2018embodied}
A.~Das, S.~Datta, G.~Gkioxari, S.~Lee, D.~Parikh, and D.~Batra, ``Embodied
  question answering,'' in \emph{Proc. IEEE Conf. Comput. Vis. Pattern
  Recognit. (CVPR)}, Jun. 2018, pp. 2054--2063.

\bibitem{deng2009imagenet}
J.~Deng, W.~Dong, R.~Socher, L.-J. Li, K.~Li, and L.~Fei-Fei, ``Imagenet: A
  large-scale hierarchical image database,'' in \emph{Proc. IEEE Conf. Comput
  vis. Pattern Recognit. (CVPR)}, Jun. 2009, pp. 248--255.

\bibitem{lin2014microsoft}
T.-Y. Lin, M.~Maire, S.~Belongie, J.~Hays, P.~Perona, D.~Ramanan,
  P.~Doll{\'a}r, and C.~L. Zitnick, ``Microsoft coco: Common objects in
  context,'' in \emph{Proc. Eur. Conf. Comput. Vis. (ECCV)}, Sep. 2014, pp.
  740--755.

\bibitem{krishna2017visual}
R.~Krishna, Y.~Zhu, O.~Groth, J.~Johnson, K.~Hata, J.~Kravitz, S.~Chen,
  Y.~Kalantidis, L.-J. Li, D.~A. Shamma \emph{et~al.}, ``Visual genome:
  Connecting language and vision using crowdsourced dense image annotations,''
  \emph{Int. J. Comput. Vis.}, vol. 123, no.~1, pp. 32--73, May. 2017.

\bibitem{chaplot2018gated}
D.~S. Chaplot, K.~M. Sathyendra, R.~K. Pasumarthi, D.~Rajagopal, and
  R.~Salakhutdinov, ``Gated-attention architectures for task-oriented language
  grounding,'' in \emph{Proc. AAAI Conf. Artif. Intell. (AAAI)}, Feb. 2018, pp.
  2819--2826.

\bibitem{wortsman2019learning}
M.~Wortsman, K.~Ehsani, M.~Rastegari, A.~Farhadi, and R.~Mottaghi, ``Learning
  to learn how to learn: Self-adaptive visual navigation using meta-learning,''
  in \emph{Proc. IEEE Int. Conf. Comput. Vis. Pattern Recognit. (CVPR)}, Jun.
  2019, pp. 6750--6759.

\bibitem{wu2018building}
\BIBentryALTinterwordspacing
Y.~Wu, Y.~Wu, G.~Gkioxari, and Y.~Tian, ``Building generalizable agents with a
  realistic and rich 3d environmentc,'' 2018. [Online]. Available:
  \url{https://arxiv.org/abs/1801.02209.}
\BIBentrySTDinterwordspacing

\bibitem{dhiman2018critical}
\BIBentryALTinterwordspacing
V.~Dhiman, S.~Banerjee, B.~Griffin, J.~M. Siskind, and J.~J. Corso, ``A
  critical investigation of deep reinforcement learning for navigation,'' 2018.
  [Online]. Available: \url{https://arxiv.org/abs/1802.02274.}
\BIBentrySTDinterwordspacing

\bibitem{oh2016control}
J.~Oh, V.~Chockalingam, H.~Lee \emph{et~al.}, ``Control of memory, active
  perception, and action in minecraft,'' in \emph{Proc. Int. Conf. Mach. Learn.
  (ICML)}, Jun. 2016, pp. 2790--2799.

\bibitem{parisotto2018neural}
E.~Parisotto and R.~Salakhutdinov, ``Neural map: Structured memory for deep
  reinforcement learning,'' in \emph{Proc. Int. Conf. Learn. Represent.
  (ICLR)}, Apr. 2018.

\bibitem{zhang2017deep}
J.~Zhang, J.~T. Springenberg, J.~Boedecker, and W.~Burgard, ``Deep
  reinforcement learning with successor features for navigation across similar
  environments,'' in \emph{Proc. IEEE Int. Conf. Intell. Rob. Syst. (IROS)},
  Sep. 2017, pp. 2371--2378.

\bibitem{mousavian2019visual}
A.~Mousavian, A.~Toshev, M.~Fi{\v{s}}er, J.~Ko{\v{s}}eck{\'a}, A.~Wahid, and
  J.~Davidson, ``Visual representations for semantic target driven
  navigation,'' in \emph{Proc. IEEE Int. Conf. Rob. Autom. (ICRA)}, May. 2019,
  pp. 8846--8852.

\bibitem{andrychowicz2017hindsight}
M.~Andrychowicz, F.~Wolski, A.~Ray, J.~Schneider, R.~Fong, P.~Welinder,
  B.~McGrew, J.~Tobin, O.~P. Abbeel, and W.~Zaremba, ``Hindsight experience
  replay,'' in \emph{Proc. Adv. Neural Inf. Process. Syst. (NIPS)}, Dec. 2017,
  pp. 5048--5058.

\bibitem{kolve2017ai2}
\BIBentryALTinterwordspacing
E.~Kolve, R.~Mottaghi, D.~Gordon, Y.~Zhu, A.~Gupta, and A.~Farhadi, ``Ai2-thor:
  An interactive 3d environment for visual ai,'' 2017. [Online]. Available:
  \url{https://arxiv.org/abs/1712.05474.}
\BIBentrySTDinterwordspacing

\bibitem{hochreiter1997long}
S.~Hochreiter and J.~Schmidhuber, ``Long short-term memory,'' \emph{Neural
  Comput.}, vol.~9, no.~8, pp. 1735--1780, Nov. 1997.

\bibitem{bonin2008visual}
F.~Bonin-Font, A.~Ortiz, and G.~Oliver, ``Visual navigation for mobile robots:
  A survey,'' \emph{J. Intell. Robotic Syst.}, vol.~53, no.~3, pp. 263--296,
  Nov. 2008.

\bibitem{gupta2017cognitive}
S.~Gupta, J.~Davidson, S.~Levine, R.~Sukthankar, and J.~Malik, ``Cognitive
  mapping and planning for visual navigation,'' in \emph{Proc. IEEE Int. Conf.
  Comput. Vis. Pattern Recognit. (CVPR)}, Jul. 2017, pp. 2616--2625.

\bibitem{savinov2018semiparametric}
N.~Savinov, A.~Dosovitskiy, and V.~Koltun, ``Semi-parametric topological memory
  for navigation,'' in \emph{Proc. Int. Conf. Learn. Represent. (ICLR)}, Apr.
  2018.

\bibitem{zhu2017visual}
Y.~Zhu, D.~Gordon, E.~Kolve, D.~Fox, L.~Fei-Fei, A.~Gupta, R.~Mottaghi, and
  A.~Farhadi, ``Visual semantic planning using deep successor
  representations,'' in \emph{Proc. IEEE Int. Conf. Comput. Vis. (ICCV)}, Oct.
  2017, pp. 483--492.

\bibitem{wu2019bayesian}
Y.~Wu, Y.~Wu, A.~Tamar, S.~Russell, G.~Gkioxari, and Y.~Tian, ``Bayesian
  relational memory for semantic visual navigation,'' in \emph{Proc. IEEE Int.
  Conf. Comput. Vis. (ICCV)}, Oct. 2019, pp. 2769--2779.

\bibitem{yang2019visual}
W.~Yang, X.~Wang, A.~Farhadi, A.~Gupta, and R.~Mottaghi, ``Visual semantic
  navigation using scene priors,'' in \emph{Proc. Int. Conf. Learn. Represent.
  (ICLR)}, May. 2019.

\bibitem{mei2016listen}
H.~Mei, M.~Bansal, and M.~R. Walter, ``Listen, attend, and walk: Neural mapping
  of navigational instructions to action sequences,'' in \emph{Proc. AAAI Conf.
  Artif. Intell. (AAAI)}, Feb. 2016, pp. 2772--2778.

\bibitem{oh2017zero}
J.~Oh, S.~Singh, H.~Lee, and P.~Kohli, ``Zero-shot task generalization with
  multi-task deep reinforcement learning,'' in \emph{Proc. Int. Conf. Mach.
  Learn. (ICML)}, Aug. 2017, pp. 2661--2670.

\bibitem{wang2018look}
X.~Wang, W.~Xiong, H.~Wang, and W.~Yang~Wang, ``Look before you leap: Bridging
  model-free and model-based reinforcement learning for planned-ahead
  vision-and-language navigation,'' in \emph{Proc. Eur. Conf. Comput. Vis.
  (ECCV)}, Sep. 2018, pp. 37--53.

\bibitem{fried2018speaker}
D.~Fried, R.~Hu, V.~Cirik, A.~Rohrbach, J.~Andreas, L.-P. Morency,
  T.~Berg-Kirkpatrick, K.~Saenko, D.~Klein, and T.~Darrell, ``Speaker-follower
  models for vision-and-language navigation,'' in \emph{Proc. Adv. Neural Inf.
  Process. Syst. (NIPS)}, Dec. 2018, pp. 3314--3325.

\bibitem{wang2019reinforced}
X.~Wang, Q.~Huang, A.~Celikyilmaz, J.~Gao, D.~Shen, Y.-F. Wang, W.~Y. Wang, and
  L.~Zhang, ``Reinforced cross-modal matching and self-supervised imitation
  learning for vision-language navigation,'' in \emph{Proc. IEEE Int. Conf.
  Comput. Vis. Pattern Recognit. (CVPR)}, Jun. 2019, pp. 6629--6638.

\bibitem{gordon2018iqa}
D.~Gordon, A.~Kembhavi, M.~Rastegari, J.~Redmon, D.~Fox, and A.~Farhadi, ``Iqa:
  Visual question answering in interactive environments,'' in \emph{Proc. IEEE
  Int. Conf. Comput. Vis. Pattern Recognit. (CVPR)}, Jun. 2018, pp. 4089--4098.

\bibitem{kempka2016vizdoom}
M.~Kempka, M.~Wydmuch, G.~Runc, J.~Toczek, and W.~Ja{\'s}kowski, ``Vizdoom: A
  doom-based ai research platform for visual reinforcement learning,'' in
  \emph{Proc. IEEE Conf. Comput. Intell. Games. (CIG)}, Sep. 2016, pp. 1--8.

\bibitem{beattie2016deepmind}
\BIBentryALTinterwordspacing
C.~Beattie, J.~Z. Leibo, D.~Teplyashin, T.~Ward, M.~Wainwright, H.~K{\"u}ttler,
  A.~Lefrancq, S.~Green, V.~Vald{\'e}s, A.~Sadik \emph{et~al.}, ``Deepmind
  lab,'' 2016. [Online]. Available: \url{https://arxiv.org/abs/1612.03801.}
\BIBentrySTDinterwordspacing

\bibitem{kulkarni2016deep}
\BIBentryALTinterwordspacing
T.~D. Kulkarni, A.~Saeedi, S.~Gautam, and S.~J. Gershman, ``Deep successor
  reinforcement learning,'' 2016. [Online]. Available:
  \url{https://arxiv.org/abs/1606.02396.}
\BIBentrySTDinterwordspacing

\bibitem{tessler2017deep}
C.~Tessler, S.~Givony, T.~Zahavy, D.~J. Mankowitz, and S.~Mannor, ``A deep
  hierarchical approach to lifelong learning in minecraft,'' in \emph{Proc.
  AAAI Conf. Artif. Intell. (AAAI)}, Feb. 2017, pp. 1553--1561.

\bibitem{mirowski2017learning}
P.~Mirowski, R.~Pascanu, F.~Viola, H.~Soyer, A.~J. Ballard, A.~Banino,
  M.~Denil, R.~Goroshin, L.~Sifre, K.~Kavukcuoglu \emph{et~al.}, ``Learning to
  navigate in complex environments,'' in \emph{Proc. Int. Conf. Learn.
  Represent. (ICLR)}, Apr. 2017.

\bibitem{jaderberg2017reinforcement}
M.~Jaderberg, V.~Mnih, W.~M. Czarnecki, T.~Schaul, J.~Z. Leibo, D.~Silver, and
  K.~Kavukcuoglu, ``Reinforcement learning with unsupervised auxiliary tasks,''
  in \emph{Proc. Int. Conf. Learn. Represent. (ICLR)}, Apr. 2017.

\bibitem{chaplot2016transfer}
D.~S. Chaplot, G.~Lample, K.~M. Sathyendra, and R.~Salakhutdinov, ``Transfer
  deep reinforcement learning in 3d environments: An empirical study,'' in
  \emph{Proc. Adv. Neural Inf. Process. Syst. (NIPS)}, Dec. 2016.

\bibitem{song2017semantic}
S.~Song, F.~Yu, A.~Zeng, A.~X. Chang, M.~Savva, and T.~Funkhouser, ``Semantic
  scene completion from a single depth image,'' in \emph{Proc. IEEE Int. Conf.
  Comput. Vis. Pattern Recognit. (CVPR)}, Jul. 2017, pp. 1746--1754.

\bibitem{chang2018matterport3d}
A.~Chang, A.~Dai, T.~A. Funkhouser, M.~Halber, M.~Niebner, M.~Savva, S.~Song,
  A.~Zeng, and Y.~Zhang, ``Matterport3d: Learning from rgb-d data in indoor
  environments,'' in \emph{Proc. IEEE Int. Conf. 3D Vis. (3DV)}, Oct. 2018, pp.
  667--676.

\bibitem{pathak2018zero}
D.~Pathak, P.~Mahmoudieh, G.~Luo, P.~Agrawal, D.~Chen, Y.~Shentu, E.~Shelhamer,
  J.~Malik, A.~A. Efros, and T.~Darrell, ``Zero-shot visual imitation,'' in
  \emph{Proc. IEEE Int. Conf. Comput. Vis. Pattern Recognit. (CVPR)}, Jun.
  2018, pp. 2050--2053.

\bibitem{mnih2015human}
V.~Mnih, K.~Kavukcuoglu, D.~Silver, A.~A. Rusu, J.~Veness, M.~G. Bellemare,
  A.~Graves, M.~Riedmiller, A.~K. Fidjeland, G.~Ostrovski \emph{et~al.},
  ``Human-level control through deep reinforcement learning,'' \emph{Nature},
  vol. 518, no. 7540, p. 529, Feb. 2015.

\bibitem{silver2016mastering}
D.~Silver, A.~Huang, C.~J. Maddison, A.~Guez, L.~Sifre, G.~Van Den~Driessche,
  J.~Schrittwieser, I.~Antonoglou, V.~Panneershelvam, M.~Lanctot \emph{et~al.},
  ``Mastering the game of go with deep neural networks and tree search,''
  \emph{nature}, vol. 529, no. 7587, p. 484, Jan. 2016.

\bibitem{lillicrap2016continuous}
T.~P. Lillicrap, J.~J. Hunt, A.~Pritzel, N.~Heess, T.~Erez, Y.~Tassa,
  D.~Silver, and D.~Wierstra, ``Continuous control with deep reinforcement
  learning,'' in \emph{Proc. Int. Conf. Learn. Represent. (ICLR)}, Apr. 2016.

\bibitem{van2016deep}
H.~Van~Hasselt, A.~Guez, and D.~Silver, ``Deep reinforcement learning with
  double q-learning,'' in \emph{Proc. AAAI Conf. Artif. Intell. (AAAI)}, Mar.
  2016, pp. 2094--2100.

\bibitem{schaul2016prioritized}
T.~Schaul, J.~Quan, I.~Antonoglou, and D.~Silver, ``Prioritized experience
  replay,'' in \emph{Proc. Int. Conf. Learn. Represent. (ICLR)}, Apr. 2016.

\bibitem{wang2016dueling}
Z.~Wang, T.~Schaul, M.~Hessel, H.~Hasselt, M.~Lanctot, and N.~Freitas,
  ``Dueling network architectures for deep reinforcement learning,'' in
  \emph{Proc. Int. Conf. Mach. Learn. (ICML)}, Jun. 2016, pp. 1995--2003.

\bibitem{mnih2016asynchronous}
V.~Mnih, A.~P. Badia, M.~Mirza, A.~Graves, T.~Lillicrap, T.~Harley, D.~Silver,
  and K.~Kavukcuoglu, ``Asynchronous methods for deep reinforcement learning,''
  in \emph{Proc. Int. Conf. Mach. Learn. (ICML)}, Jun. 2016, pp. 1928--1937.

\bibitem{schulman2017proximal}
\BIBentryALTinterwordspacing
J.~Schulman, F.~Wolski, P.~Dhariwal, A.~Radford, and O.~Klimov, ``Proximal
  policy optimization algorithms,'' 2017. [Online]. Available:
  \url{https://arxiv.org/abs/1707.06347.}
\BIBentrySTDinterwordspacing

\bibitem{niepert2016learning}
M.~Niepert, M.~Ahmed, and K.~Kutzkov, ``Learning convolutional neural networks
  for graphs,'' in \emph{Proc. Int. Conf. Mach. Learn. (ICML)}, Jun. 2016, pp.
  2014--2023.

\bibitem{marino2017more}
K.~Marino, R.~Salakhutdinov, and A.~Gupta, ``The more you know: Using knowledge
  graphs for image classification,'' in \emph{Proc. IEEE Int. Conf. Comput.
  Vis. Pattern Recognit. (CVPR)}, Jul. 2017, pp. 2673--2681.

\bibitem{wang2018zero}
X.~Wang, Y.~Ye, and A.~Gupta, ``Zero-shot recognition via semantic embeddings
  and knowledge graphs,'' in \emph{Proc. IEEE Int. Conf. Comput. Vis. Pattern
  Recognit. (CVPR)}, Jun. 2018, pp. 6857--6866.

\bibitem{norcliffe2018learning}
W.~Norcliffe-Brown, S.~Vafeias, and S.~Parisot, ``Learning conditioned graph
  structures for interpretable visual question answering,'' in \emph{Proc. Adv.
  Neural Inf. Process. Syst. (NIPS)}, Dec. 2018, pp. 8334--8343.

\bibitem{hu2018relation}
H.~Hu, J.~Gu, Z.~Zhang, J.~Dai, and Y.~Wei, ``Relation networks for object
  detection,'' in \emph{Proc. IEEE Int. Conf. Comput. Vis. Pattern Recognit.
  (CVPR)}, Jun. 2018, pp. 3588--3597.

\bibitem{wang2018deep}
Z.~Wang, T.~Chen, J.~Ren, W.~Yu, H.~Cheng, and L.~Lin, ``Deep reasoning with
  knowledge graph for social relationship understanding,'' in \emph{Proc. Joint
  Conf. Artif. Intell. (IJCAI)}, Jul. 2018, pp. 1021--1028.

\bibitem{gao2019know}
J.~Gao, T.~Zhang, and C.~Xu, ``I know the relationships: Zero-shot action
  recognition via two-stream graph convolutional networks and knowledge
  graphs,'' in \emph{Proc. AAAI Conf. Artif. Intell. (AAAI)}, Feb. 2019, pp.
  8303--8311.

\bibitem{he2016deep}
K.~He, X.~Zhang, S.~Ren, and J.~Sun, ``Deep residual learning for image
  recognition,'' in \emph{Proc. IEEE Int. Conf. Comput. Vis. Pattern Recognit.
  (CVPR)}, Jun. 2016, pp. 770--778.

\bibitem{chopra2005learning}
S.~Chopra, R.~Hadsell, Y.~LeCun \emph{et~al.}, ``Learning a similarity metric
  discriminatively, with application to face verification,'' in \emph{Proc.
  IEEE Int. Conf. Comput. Vis. Pattern Recognit. (CVPR)}, Jun. 2005, pp.
  539--546.

\bibitem{kipf2017semi}
T.~N. Kipf and M.~Welling, ``Semi-supervised classification with graph
  convolutional networks,'' in \emph{Proc. Int. Conf. Learn. Represent.
  (ICLR)}, Apr. 2017.

\bibitem{redmon2018yolov3}
\BIBentryALTinterwordspacing
J.~Redmon and A.~Farhadi, ``Yolov3: An incremental improvement,'' 2018.
  [Online]. Available: \url{https://arxiv.org/abs/1804.02767.}
\BIBentrySTDinterwordspacing

\bibitem{anderson2018evaluation}
\BIBentryALTinterwordspacing
P.~Anderson, A.~Chang, D.~S. Chaplot, A.~Dosovitskiy, S.~Gupta, V.~Koltun,
  J.~Kosecka, J.~Malik, R.~Mottaghi, M.~Savva \emph{et~al.}, ``On evaluation of
  embodied navigation agents,'' 2018. [Online]. Available:
  \url{https://arxiv.org/abs/1807.06757.}
\BIBentrySTDinterwordspacing

\bibitem{abadi2016tensorflow}
\BIBentryALTinterwordspacing
M.~Abadi, A.~Agarwal, P.~Barham, E.~Brevdo, Z.~Chen, C.~Citro, G.~S. Corrado,
  A.~Davis, J.~Dean, M.~Devin \emph{et~al.}, ``Tensorflow: Large-scale machine
  learning on heterogeneous distributed systems,'' 2016. [Online]. Available:
  \url{https://arxiv.org/abs/1603.04467.}
\BIBentrySTDinterwordspacing

\bibitem{maaten2008visualizing}
L.~v.~d. Maaten and G.~Hinton, ``Visualizing data using t-sne,'' \emph{J. Mach.
  Learn. Res.}, vol.~9, no. Nov, pp. 2579--2605, Nov. 2008.

\end{thebibliography}

\end{document}